\documentclass{llncs}

\usepackage{graphicx}
\usepackage{subfigure}
\usepackage{array}

\usepackage{amsmath}
\usepackage{amssymb}
\usepackage{bm}

\usepackage{cite}

\usepackage{url}
\urldef{\mailsa}\path|boris.oreshkin@mail.mcgill.ca|
\urldef{\mailsb}\path|arbel@cim.mcgill.ca|

\graphicspath{{fig/}}

\newcommand{\bX}{{\mathbf X}}

\newcommand{\bq}{{\mathbf q}}
\newcommand{\bu}{{\mathbf u}}
\newcommand{\br}{{\mathbf r}}
\newcommand{\bx}{{\mathbf x}}

\newcommand{\opt}{\mathrm{opt}}

\begin{document}

\pagestyle{headings}

\mainmatter              

\title{Optimization over Random and Gradient Probabilistic Pixel Sampling for Fast, Robust Multi-Resolution Image Registration}

\titlerunning{Optimization over Random and Gradient Probabilistic Pixel Sampling}

\author{Boris N. Oreshkin \and Tal Arbel}

\authorrunning{Boris N. Oreshkin \and Tal Arbel} 

\tocauthor{ }

\institute{McGill University, Center of Intelligent Machines,\\
3480 University Street, Montreal, Quebec, Canada, H3A 2A7\\
\mailsa, \mailsb}

\maketitle              

\begin{abstract}
This paper presents an approach to fast image registration through probabilistic pixel sampling. We propose a practical scheme to leverage the benefits of two state-of-the-art pixel sampling approaches: gradient magnitude based pixel sampling and uniformly random sampling. Our framework involves learning the optimal balance between the two sampling schemes off-line during training, based on a small training dataset, using particle swarm optimization. We then test the proposed sampling approach on 3D rigid registration against two state-of-the-art approaches based on the popular, publicly available, Vanderbilt RIRE dataset. Our results indicate that the proposed sampling approach yields much faster, accurate and robust registration results when compared against the state-of-the-art.

\end{abstract}

\section{Introduction} \label{sec:intro}

Image registration is one of the critical problems in the field of medical imaging. It transcends wide range of applications from image-guided interventions to building anatomical atlases from patient data. Typically, the evaluation of the similarity measure and its derivatives are required to perform the optimization over transformation parameters. However, performing these computations based on all the available image pixels can be prohibitively costly. The expense is mainly due to the large number of pixel intensity values involved in the calculations. Time-sensitive applications, like image guided intervention, generally benefit from techniques to speed up direct image registration by utilizing only a subset of available pixels during registration. In these contexts, several percent of accuracy decrease could be tolerated and traded for preservation of robustness and significant decrease in registration time. However, significant speedups attained via \emph{aggressive} reduction in the number of selected pixels (less than $1\%$ of the total number of pixels) often result in deterioration of robustness (increase in failure rate) and relatively rapid increase of registration error.

Many pixel sampling schemes have been suggested in the literature. Uniformly random pixel selection (URS), in which a random subset of all pixels sampled with uniform probabilities is used to drive the optimization, gained popularity due to its simplicity and robustness~\cite{Mattes2003,Viola1997}. Other techniques strived to improve registration accuracy by optimizing the pixel selection process. The deterministic pixel selection strategy~\cite{Reeves1995} consists in calculating a selection criterion for each pixel (e.g. based on Jacobian of the cost function~\cite{Dellaert1999}) and comparing it to the threshold. The subset of pixels whose selection criterion values transcend a predefined threshold are used for registration. This led to a  clustering phenomenon, as pointed out by Dallaert and Collins~\cite{Dellaert1999}, who attempted to overcome this effect and proposed a probabilistic pixel selection strategy that uniformly samples from subset of pixels having top twenty percent values of selection criterion pixels. Brooks and Arbel~\cite{Brooks2007} extend the approach of Dellaert and Collins~\cite{Dellaert1999} by proposing an information theoretic selection criterion and by addressing the issue of Jacobian scale inherent to the gradient descent type optimization algorithms. Benhimane et al.~\cite{Benhimane2007} proposed a criterion to speed up the convergence of the optimization by selecting only the pixels that closely verify the approximation made by the optimization. Sabuncu and Ramadge used information theoretical approach to demonstrate the fact that the pixel sampling scheme should emphasize pixels with high spatial gradient magnitude~\cite{Sabuncu2004}. Here the moving image is probabilistically subsampled using non-uniform grid generated based on the probabilities proportional to the gradient magnitude. This approach allows to diversify and spread subsampled pixels while still giving attention to image details. This approach alleviates the effects of selected pixel clustering inherent to deterministic pixel selection strategy discussed e.g. by Reeves and Hezar~\cite{Reeves1995} while still allowing to focus on the more useful pixels. Finally, curvlet based sampling, recently proposed by Freiman et al.~\cite{Freiman2010} tested on Vanderbilt RIRE dataset~\cite{Fitzpatrick1998} revealed approximately the same level of accuracy as the gradient subsampling approach~\cite{Sabuncu2004}.

Exploring the method of Sabuncu and Ramadge, one notices that the strategy works well for relatively large pixel sampling rates ($1$ to $10\%$). However, as the number of selected pixels decreases, it tends to concentrate exclusively on pixels with the highest gradient magnitude, which limits its exploratory capability and leads to deterioration of robustness and accuracy. The uniformly random sampling strategy, on the other hand, has very good exploratory behaviour as any pixel has equal probability to be used in the similarity metric calculations. At the same time, the uniformly random sampling lacks attention to image structural details that often aid in achieving easier and more accurate registration results. Thus the URS often provides better robustness, but fails to produce the same accuracy levels as the gradient magnitude based approach.

In this paper, we propose to combine the virtues of the two techniques to obtain faster and more robust image registration. We introduce a new  multi-scale sampling scheme, whereby the sampling probabilities are based on the convex combination of the uniformly random sampling probabilities~\cite{Mattes2003,Viola1997} and the gradient based sampling probabilities~\cite{Sabuncu2004}. We further propose to learn the value of the convex combination parameter off-line by optimizing the empirical target registration error obtained from a small training dataset via particle swarm optimization~\cite{Kennedy1995}. Our approach effectively serves to improve the performance of one of the best existing state-of-the-art sampling methods and achieve the greatest reduction in the number of pixels used for the evaluation of the similarity metric under the constraint of preserving the accuracy and robustness at reasonable levels. We test the proposed approach on the Vanderbilt RIRE dataset~\cite{Fitzpatrick1998}. Our results indicate that the proposed approach allows to significantly reduce the number of pixels used in the evaluation of the similarity metric and hence accelerate the registration procedure while improving robustness and preserving accuracy of the gradient based sampling technique.

\section{Problem Statement}

The direct image registration problem can be formulated for the reference $I(\bx)$ and the moving $J(\mathrm{T}_{\theta}(\bx))$ images defined by their pixel intensity values $I_i, J_i : \mathcal{X} \rightarrow \mathcal{I}, i = 1 \ldots N$ seen as mappings from the coordinate space $\mathcal{X} \subseteq \mathbb{R}^d$ to the intensity space $\mathcal{I} \subseteq \mathbb{R}$, where $d$ is the dimensionality of coordinate space and $N$ is the number of pixels (here we assume, without loss of generality, that the number of pixels in the images is equal). The problem is solved by finding the
parameters $\theta \in \Theta$ of the warp $\mathrm{T}_{\theta} : \mathcal{X} \rightarrow \mathcal{X}$ that maximize the similarity metric $D_N : \mathcal{I}^{N\times 2} \rightarrow \mathbb{R}$ that maps $N$ intensity values of the reference and $N$ intensity values of the moving images
into a number characterizing the degree of similarity between these images for a given value of the warp parameters:
\begin{align} \label{eqn:reg_problem_N}
\theta_{\opt} = \arg\max_{\theta \in \Theta} D_N[I(\bx),
J(\mathrm{T}_{\theta}(\bx))].
\end{align}
Widely used similarity metrics are mutual information~\cite{Viola1997} and normalized mutual information (NMI)~\cite{Studholme99}. The pixel selection process can be viewed as the approximate solution using the calculation of the similarity metric based on only $M$ pixels of each of the images:
\begin{align} \label{eqn:reg_problem_M}
\theta_{\opt} = \arg\max_{\theta \in \Theta}
D_M[I(\bx), J(\mathrm{T}_{\theta}(\bx))],
\end{align}
Since this solution is based on $M < N$ pixels it is less computationally expensive. As was indicated in Section~\ref{sec:intro}, the deterioration of robustness and accuracy of the existing pixel subsampling methods, and gradient based sampling in particular, is a major problem when the number of pixels used to calculate the similarity metric is small, $M \ll N$. At the same time, the small sampling rate condition $M/N \ll 1$ ensures that significant computational gain results from the pixel selection. In this paper we strive to solve the problem of robustness and accuracy deterioration for small $M$. To this end, we propose the approach to combine the uniformly random sampling with the gradient based sampling within the multi-scale framework that we discuss in detail in the next section.

\section{Proposed Algorithm} \label{sec:proposed_algorithm}

Sabuncu and Ramadge used information theoretical approach to demonstrate the fact that the pixel sampling scheme should emphasize pixels with high spatial gradient magnitude~\cite{Sabuncu2004}. Based on this observation they proposed the sampling strategy where pixel $i$ is sampled with the probability $q_i = \alpha \| \nabla J_i \|_2 $, where $\| \nabla J_i \|_2$ is the magnitude of spatial intensity gradient of pixel $i$ and $\alpha$ is the normalization factor that determines the average number of subsampled pixels. The URS sampling approach attaches equal sampling probability to each pixel. The gradient magnitude based sampling puts more emphasis on the image gradient details that provide for more accurate registration. However, this often reduces registration robustness by reducing image exploration. URS explores images well via extensive uniform sampling, but lack of attention to image details reduces its accuracy.

We propose to combine the positive properties of the two techniques just described and to obtain a better multi-scale sampling scheme designed for fast image registration. In our algorithm we combine the probabilities of the gradient magnitude based sampling approach and the URS approach such that the sampling probability of the proposed algorithm is the convex combination  of the probabilities defined by the two corresponding component approaches. The optimal value of the convex combination parameter is learned off-line by optimizing the empirical target registration error (ETRE) obtained from a training dataset. In the remainder of this section we describe the details of the proposed algorithm.

Assume that there are $R$ resolution levels in the registration scheme, $r$ is the resolution level number and $N_r$ is the number of pixels at level $r$. Assume that $r=1$ corresponds to the highest resolution level (original images) and hence $N_1 = N$. Denote $\bq^r = [q_1^r, \ldots, q_{N_r}^r]$ the vector of sampling probabilities for the gradient magnitude sampling method at level $r$ and assume that the normalization factor $\alpha^r$ at this level is chosen so that the average number of pixels equals $M_r$. Similarly, for the URS method the vector of sampling probabilities is $\bu^r = [M_r/N_r, \ldots, M_r/N_r]$ resulting in the average number of pixels sampled being equal to $M_r$. The vector of sampling probabilities for the proposed approach, $\br^r$, is the convex combination of the two previously defined vectors:
\begin{align}
\br^r = (1-\beta^r) \bq^r + \beta^r \bu^r,
\end{align}
where $\beta^r \in [0, 1]$ is the mixing parameter. Greater values of $\beta^r$ emphasize the exploration brought about by the URS and lower values of this parameter emphasize prominent image features that could facilitate more accurate registration. In a general situation we expect that at every level $r$ and for every pixel sampling rate $M_r / N_r$ there is an optimal value of $\beta^r$ that compromises image exploration and exploitation of prominent image features.

It is hard (if at all possible) to analytically formulate and solve the problem of optimizing $\beta^r$ based on statistical models of the images. At the same time, if a small, but representative training dataset is available for the registration problem at hand, the value of this parameter could be learned empirically off-line. The learned value of this parameter could then be used upon each subsequent application of the proposed algorithm. The proposed algorithm could be retrained whenever there is a need to solve a new registration problem with significantly different image specifics. This is not an unreasonable assumption since in the application domain the specifics of particular registration problem often affect e.g. the choice of similarity metric, optimization strategy and interpolation scheme. This implies that at least some training information in the form of the small set of exemplar image pairs from the problem-specific modalities using certain acquisition and post-processing protocols must always be available to the registration algorithm designer to guide the algorithm development.

Based on the assumption that we have a training data set and the gold standard registration parameters for the image pairs in this dataset we formulate the empirical learning criterion $Q^r(\beta^r)$. We define the ETRE as the average over $V$ image pairs in the training dataset and $U$ Monte-Carlo trials:
\begin{align}
Q^r(\beta^r) = \frac{1}{V}\frac{1}{U} \sum_{v=1}^{V} \sum_{u=1}^U \| \bX_v - \widehat\bX^r_{u,v}(\beta^r) \|_2^2.
\end{align}
Here $\bX_v$ is the set of transformed coordinates obtained using gold standard registration parameters for image pair $v$ and $\widehat\bX^r_{u,v}(\beta^r)$ is the set of transformed coordinates for image pair $v$ and Monte-Carlo trial $u$ found using the empirical estimate of the registration parameters obtained via the optimization of the similarity metric at resolution scale $r$ using the proposed pixel sampling algorithm with a given value of mixing parameter $\beta^r$. As the pixel sampling algorithm is randomized, some degree of Monte-Carlo averaging could be beneficial if $V$ is relatively small ($3\ldots5$ images). Thus we repeat the registration procedure for the same candidate value $\beta^r$, level $r$ and image pair $v$ $U$ times and calculate $\widehat\bX^r_{u,v}(\beta^r)$ based on the new registration parameter estimate each time.

We propose to learn the value of $\beta^r$ by minimizing the ETRE $Q^r(\beta^r)$:
\begin{align}
\widehat \beta^r = \arg\min_{\beta^r \in [0; 1]} Q^r(\beta^r).
\end{align}
The function $Q^r(\cdot)$ is generally extremely irregular and non-smooth, because of the possible registration failures and because of complex dependence of the ETRE on the value of $\beta^r$. At the same time, the domain of this function is well defined and restricted. Thus any optimizer capable of performing global or quasi-global search on a restricted interval using only the objective function values will suffice to solve this problem. We propose to use the particle swarm optimization (PSO)~\cite{Kennedy1995} in order to find $\widehat \beta^r$. Our algorithm proceeds by finding $\widehat\beta^R$, the value of the mixing parameter for the scale with the lowest resolution using PSO. The multi-scale registration algorithm proceeds from the lowest resolution level to the highest resolution level sequentially utilizing the registration parameters obtained at the lower resolution level as an initialization for the current resolution level. Our learning algorithm thus uses the identified value of $\widehat\beta^R$ to find the estimate of the registration parameters at resolution level $R$. Then the optimal value $\widehat\beta^{R-1}$ for the next higher-resolution level is found using the registration parameters identified at level $R$ as initialization. This procedure iterates until the values of mixing parameter for all resolution levels $R, R-1, \ldots, 1$ are identified.

\section{Experiments with the RIRE Vanderbilt Dataset}

\subsection{Dataset Description}

To test the proposed algorithm we made use of the real clinical data available in RIRE Vanderbilt dataset~\cite{Fitzpatrick1998}. The performance of algorithms was evaluated by registering 3D volumes corresponding to CT images to geometrically corrected MR images. MR image set included images acquired using T1, T2 and PD acquisition protocols. The total number of different image pairs used was 19. Those pairs were taken from patients 001, 002, 003, 004, 005, 006, 007 for which geometrically corrected images are available. Patients 003 and 006 did not have geometrically corrected PD and MR-T1 images respectively. According to the data exchange protocol established by the RIRE Vanderbilt project, registration results obtained via algorithms under the test were uploaded to the RIRE Vanderbilt web-site. Algorithm evaluation results were calculated by the RIRE Vanderbilt remote computer using the gold standard transformation not available to us and published on their web-site in the form of tables containing registration errors calculated over 6 to 10 volumes of interest (VOIs) for each image pair. For patient 000 geometrically corrected MR-T1, MR-T2, MR-PD images and corresponding CT image are available along with the set of transformed coordinates obtained using gold standard registration parameters. Three image pairs from patient 000 were used to learn the values of mixing parameters according to the algorithm described in Section~\ref{sec:proposed_algorithm}.

\subsection{Experimental Setup}

All images were first resampled to a common 1mm grid using bicubic interpolation. We used 4-scale registration based on the low-pass filtered and downsampled image pyramid. Resolution level number four had grid spacing 4 mm along each axis and resolution level number one had grid spacing 1mm along each axis. The estimate of the registration parameters obtained at a lower resolution level was used as a starting point for the registration at the next higher resolution level; level 4 had all its parameters initialized to zero values. We concentrated on recovering 6 rigid registration parameters (3 translations and 3 rotations) using the NMI similarity metric~\cite{Studholme99}. Histogram for the evaluation of the similarity metric was calculated using the partial volume approach with Hanning windowed sinc kernel function~\cite{Lu2008}. Similarity metric was optimized using the trust region Gauss-Newton approach~\cite{Brooks2008}. We implemented the most calculation intensive part of the code (calculation of the cost function and its derivatives) in C and benchmarked the algorithms within the MATLAB environment.

\begin{figure*}[t]
\centering
      \includegraphics[width=0.32\textwidth]{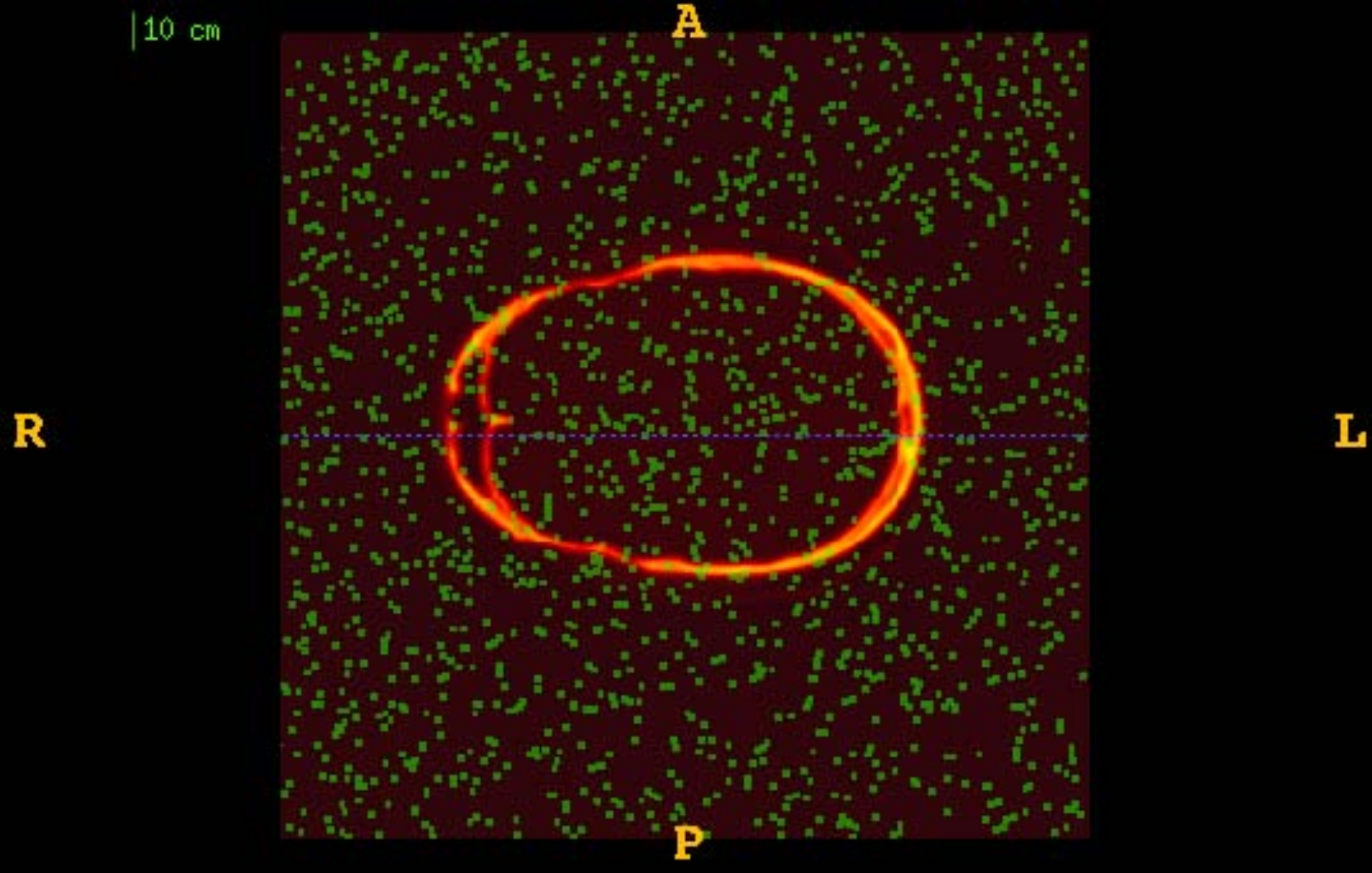}
      \includegraphics[width=0.32\textwidth]{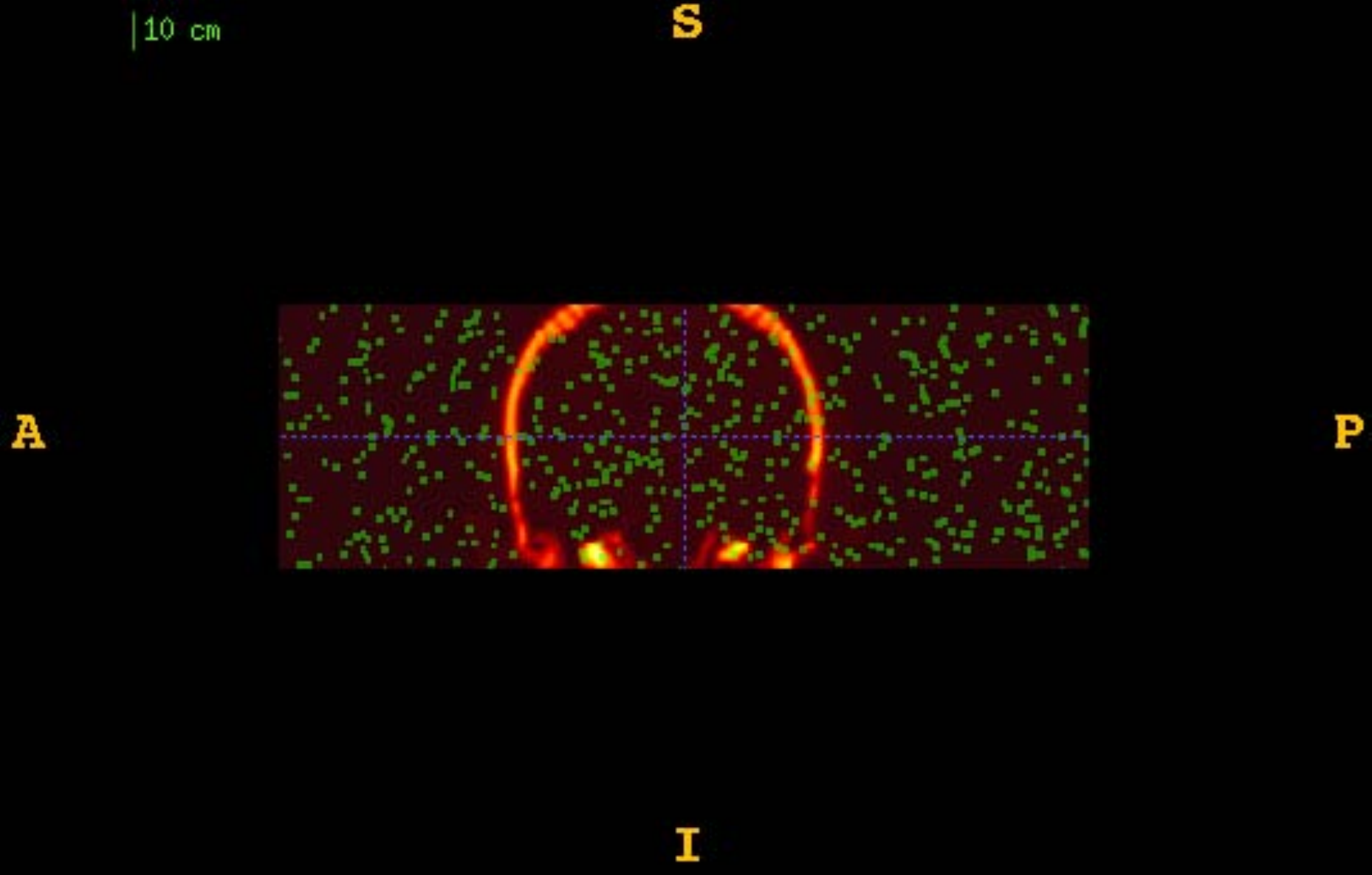}
      \includegraphics[width=0.32\textwidth]{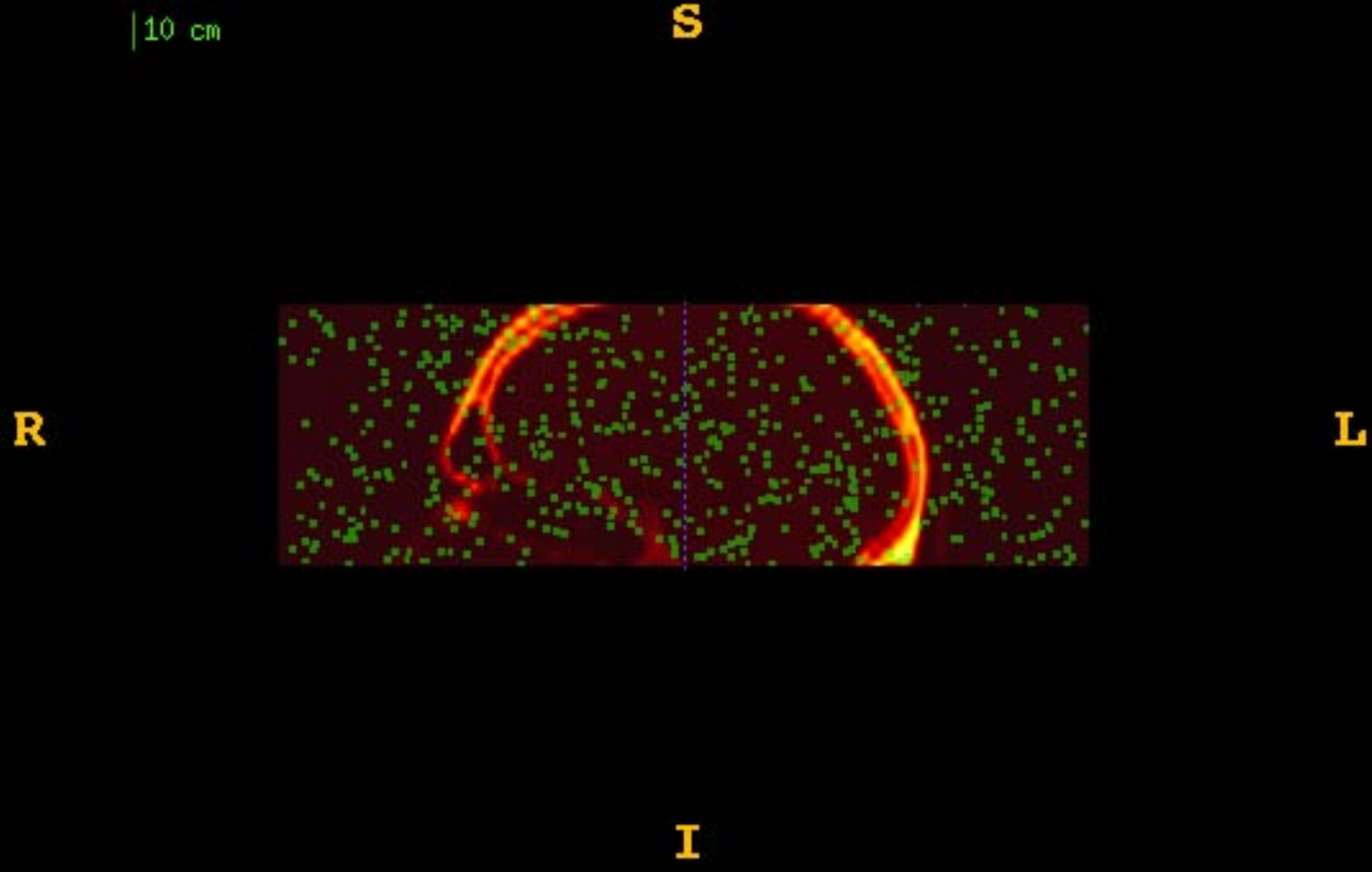}
\\
      \includegraphics[width=0.32\textwidth]{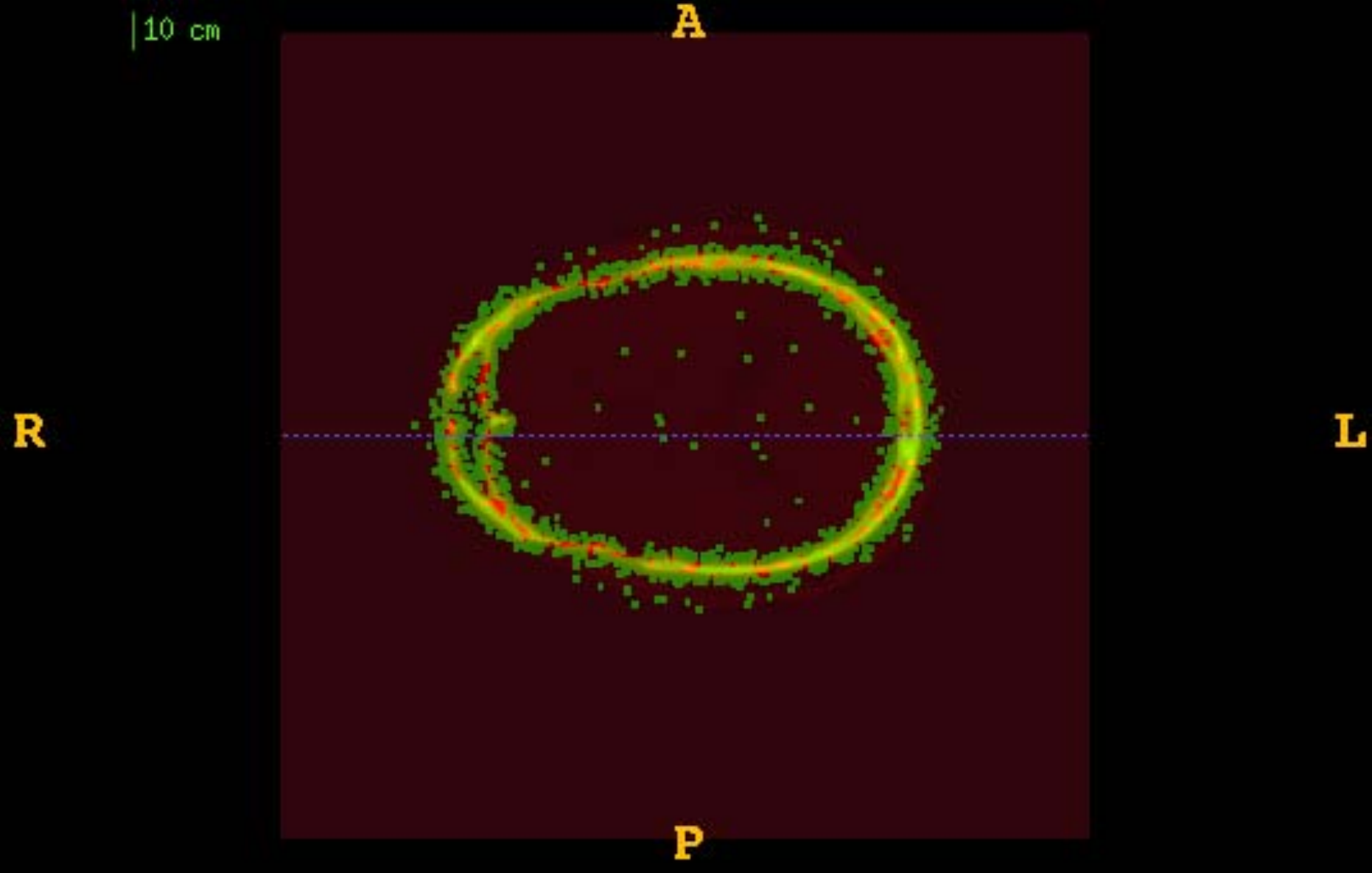}
      \includegraphics[width=0.32\textwidth]{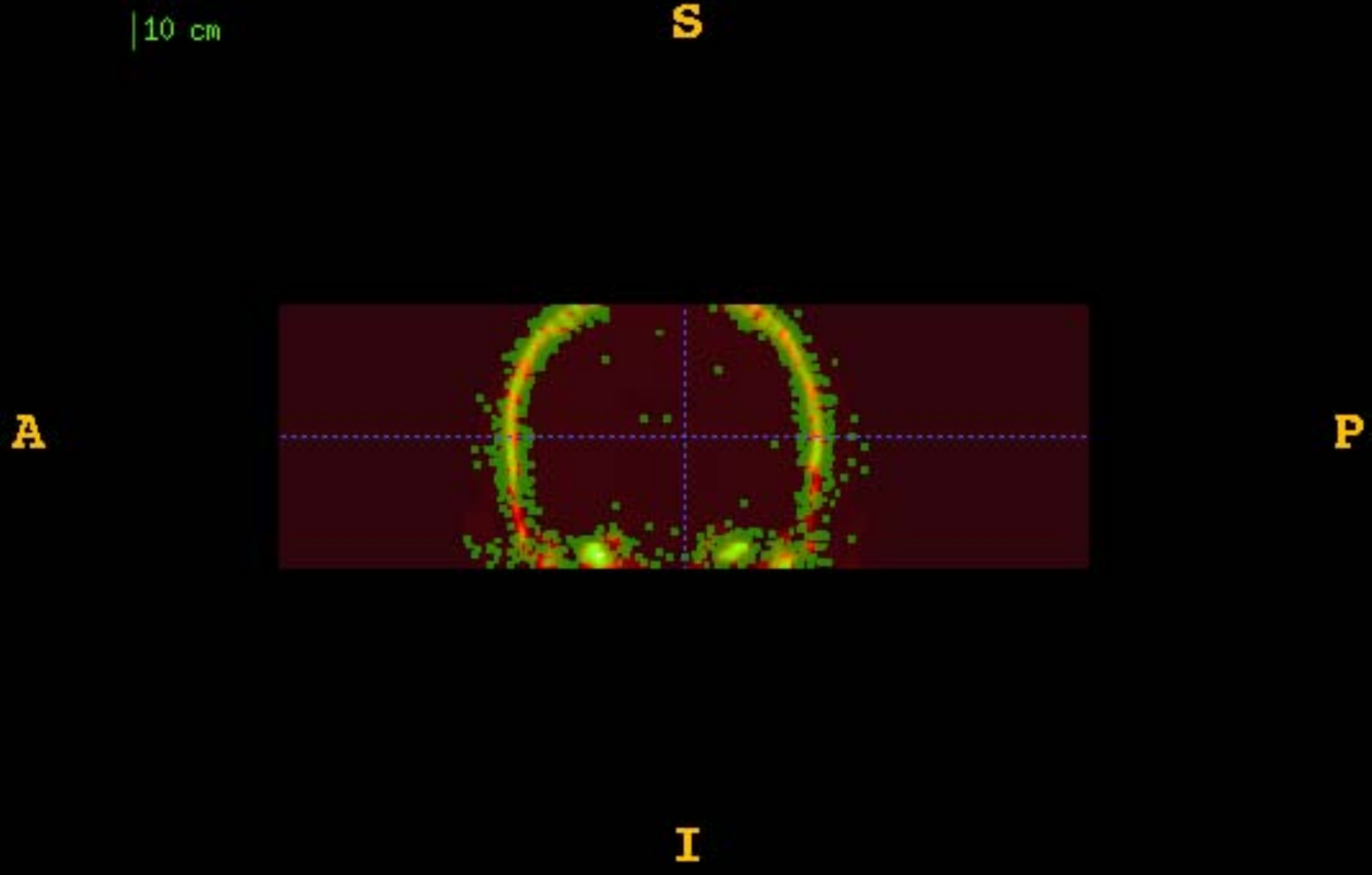}
      \includegraphics[width=0.32\textwidth]{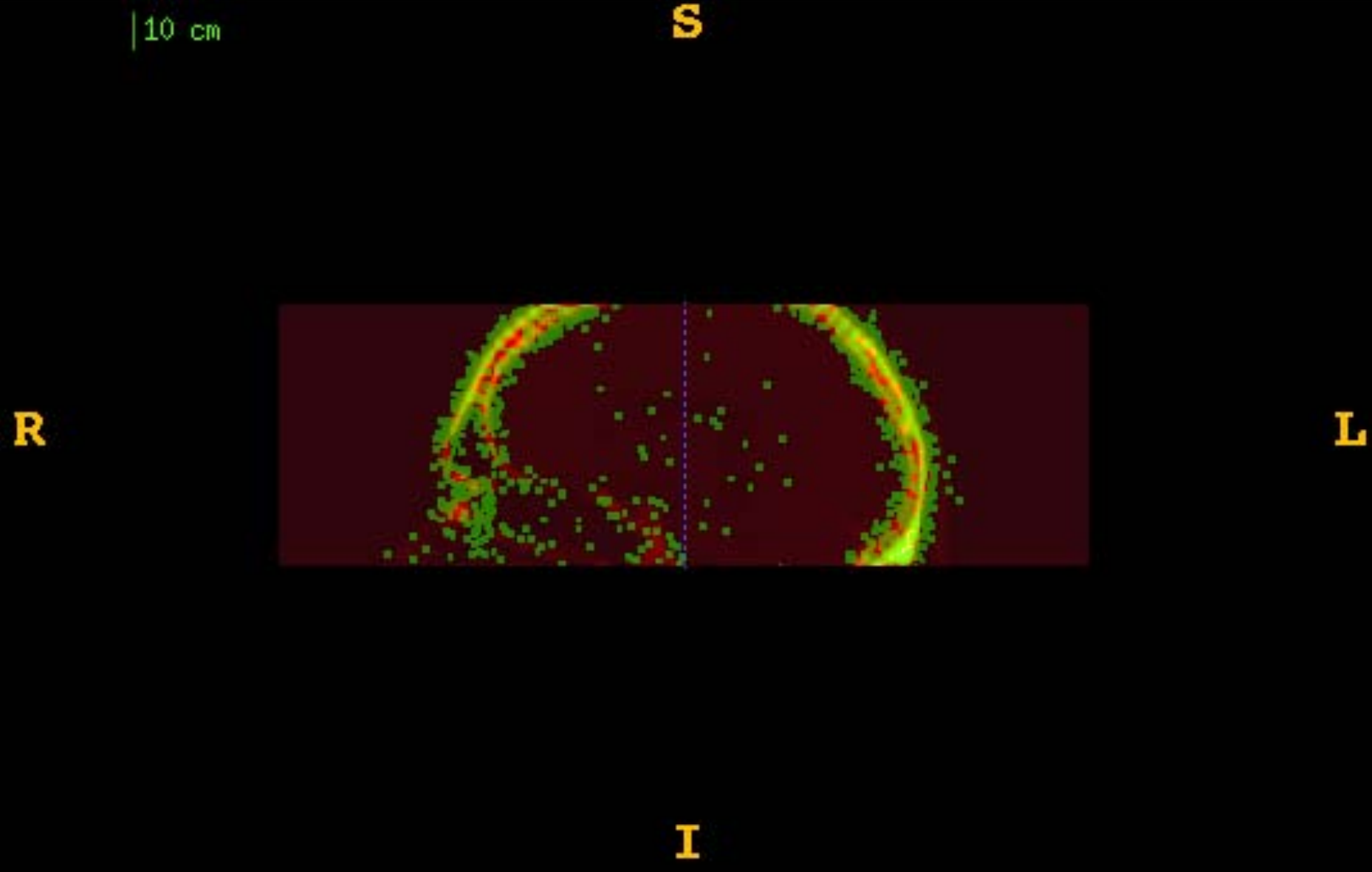}
\\
      \includegraphics[width=0.32\textwidth]{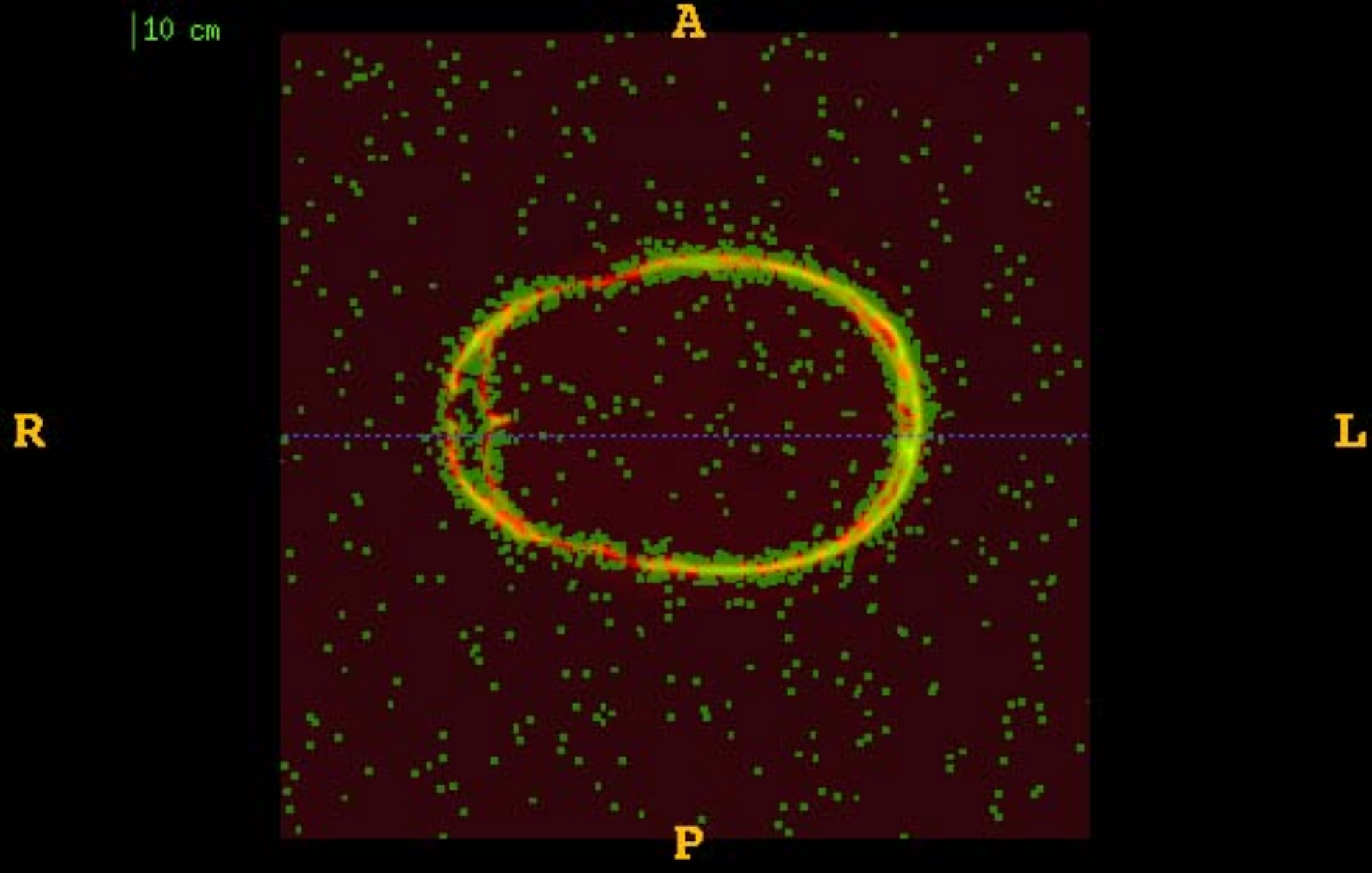}
      \includegraphics[width=0.32\textwidth]{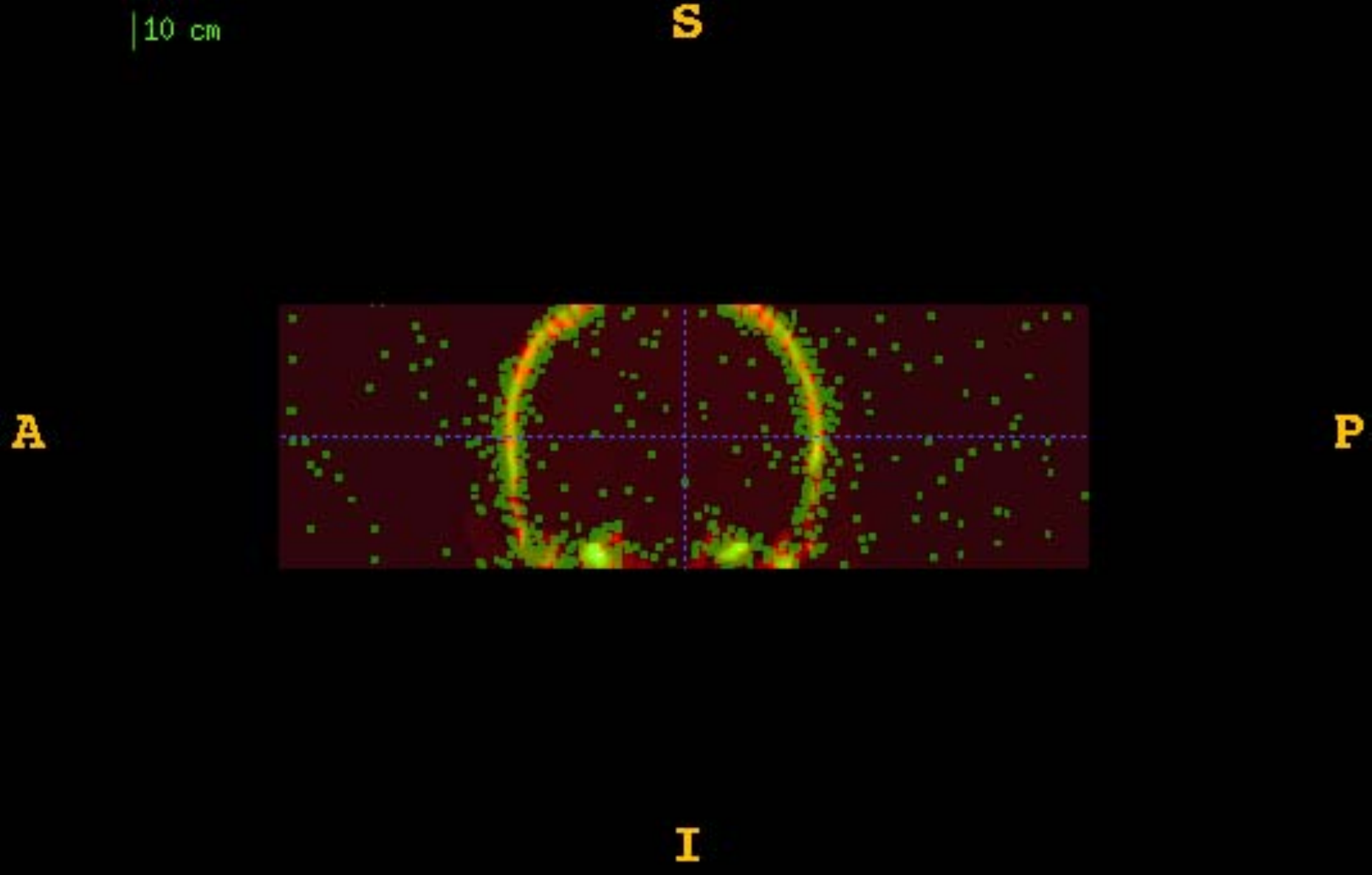}
      \includegraphics[width=0.32\textwidth]{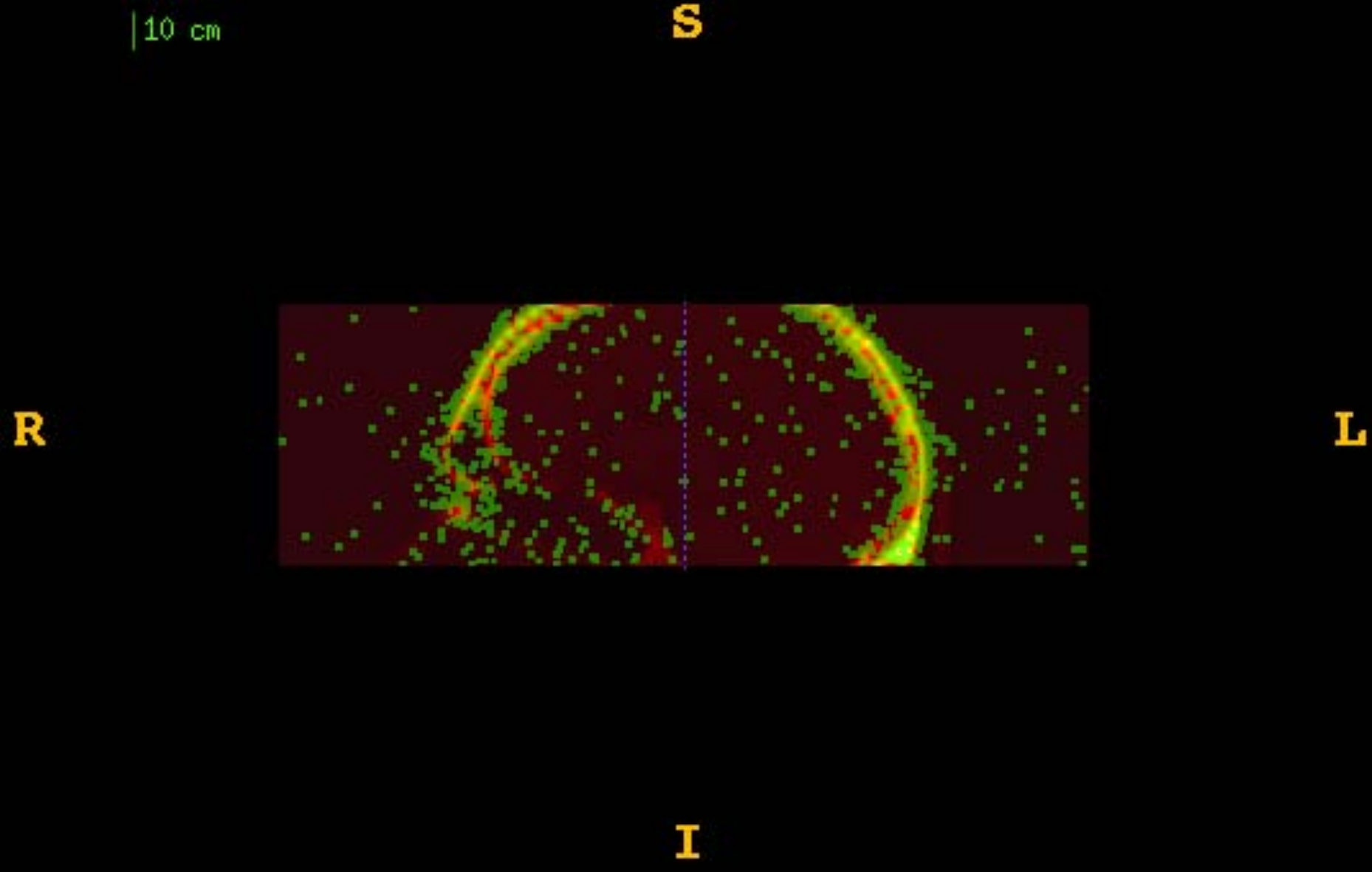}
\caption{Pixel selection masks generated using different approaches at the highest resolution level for sampling rate 0.5\%. (FIRST ROW) \textbf{URS}; (SECOND ROW) \textbf{GMS}; (THIRD ROW) proposed approach with learned value of the mixing parameter $\widehat\beta^1 = 0.2$. First column axial slice, second column sagittal slice, third column coronal slice. All images are obtained using ITK-SNAP~\cite{py06nimg}.}
\label{fig:mask_examples} 
\vspace{-0.5cm}
\end{figure*}

We evaluated the performance of following pixel sampling approaches. The uniformly random sampling (\textbf{URS}) technique consists of randomly selecting pixels with equal probabilities at every iteration~\cite{Viola1997}. At a given resolution level $r$ all pixels have equal probability of being selected, $M / N_r$ if $M < N_r$ and $1$ if $M \geq N_r$; the average number of selected pixels is thus equal to $M$ at each resolution level. Note that we used equal number of selected pixels for all resolution scales. Gradient magnitude sampling (\textbf{GMS}), a slight modification of gradient based subsampling originally proposed by Subuncu and Ramadge~\cite{Sabuncu2004}, consists in calculating spatial gradient magnitude $\|\nabla J_i\|_2 = \sqrt{(\partial J_i/\partial x_i)^2 + (\partial J_i/\partial y_i)^2 + (\partial J_i/\partial z_i)^2}$ and sampling pixels at every optimization iteration according to the probabilities proportional to it, where the proportionality coefficient is chosen so that the average number of pixels selected at every resolution scale is equal to $M$. Proposed method described in Section~\ref{sec:proposed_algorithm} (\textbf{Proposed}) consists of mixing the probabilities obtained from URS and GMS methods and learning the value of the mixing parameter using the training dataset constructed from image pairs of patient 000. We evaluate these three algorithms for the following values of pixel sampling rates (given in \%): $M/N \in \{ 0.02, 0.04, 0.065, 0.1, 0.5, 1 \}$ (sampling rate is calculated with respect to the image size at the highest resolution level, $N$).

\subsection{Results} \label{ssec:Results}

Figure~\ref{fig:mask_examples} shows the examples of pixel selection masks generated using tested approaches at the highest resolution level for pixel sampling rate 0.5\%. It is obvious that the samples generated with the URS approach are extremely spread, whereas the samples generated with the GMS approach are overly concentrated along the gradient magnitude structures present in the image. The proposed approach produces samples that balance those two extremities.

\begin{figure*}[t]
\centering
      \includegraphics[width=0.70\textwidth]{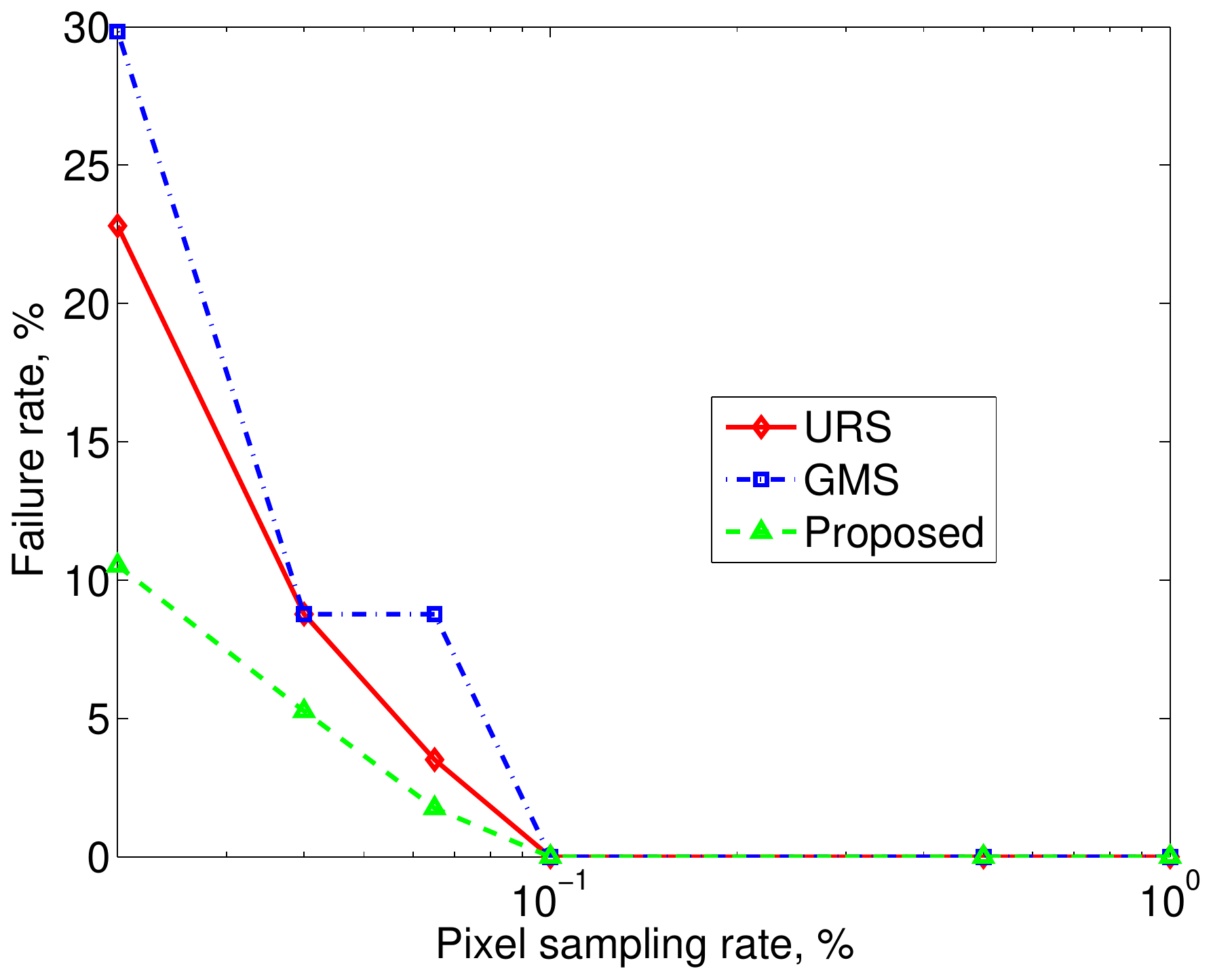}
\caption{Failure rate for different pixel sampling mechanisms: gradient magnitude sampling (GMS), uniformly random sampling (URS), Proposed. Note that the proposed approach consistently outperforms in terms of robustness.}
\label{fig:performancePlots:FailureRate} 
\vspace{-0.5cm}
\end{figure*}

\begin{figure*}[t]
\centering
      \includegraphics[width=0.70\textwidth]{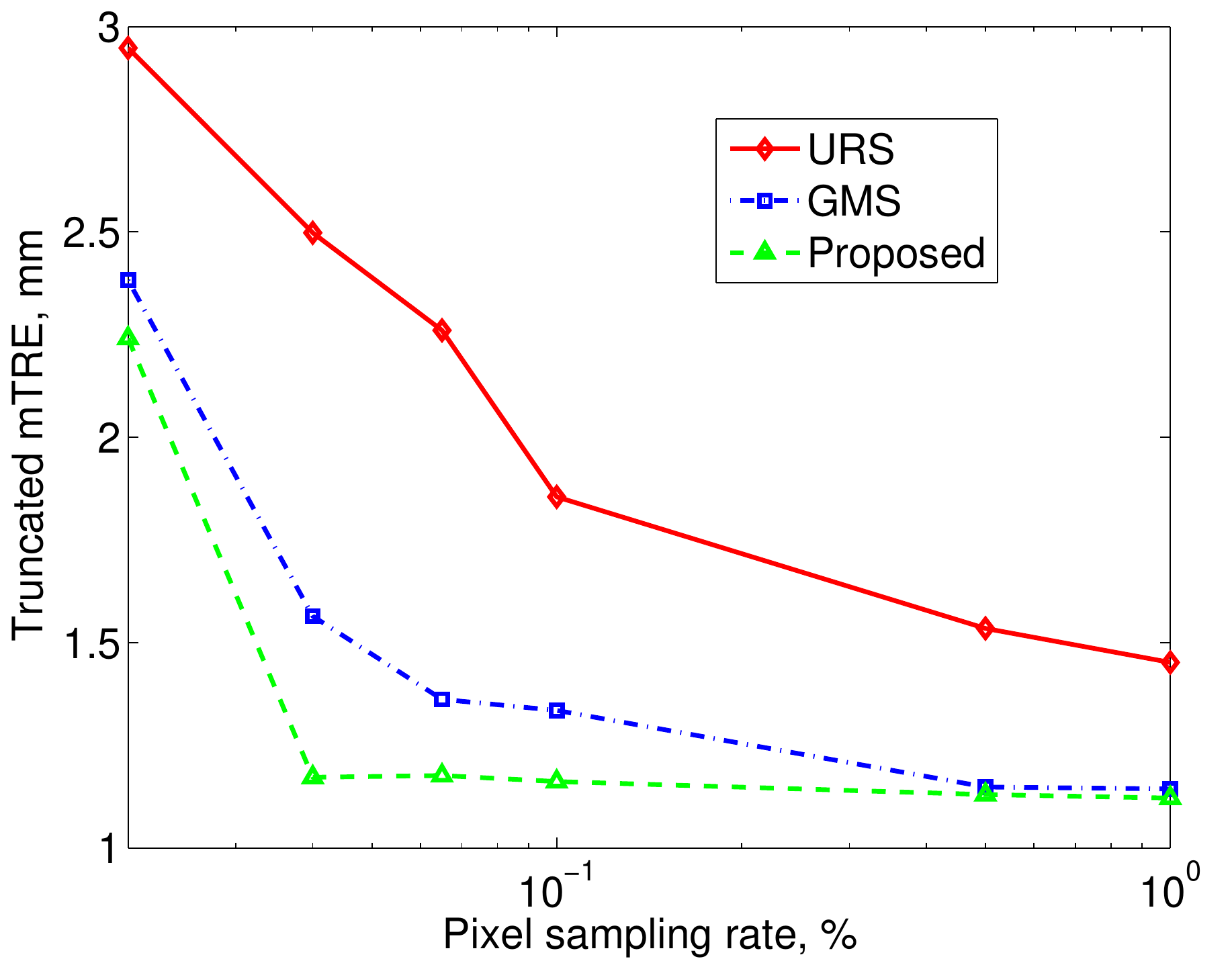}
\caption{Trimmed average registration error for different pixel sampling mechanisms: gradient magnitude sampling (GMS), uniformly random sampling (URS), Proposed. Note that the proposed approach consistently outperforms in terms of accuracy.}
\label{fig:performancePlots:mTRE} 
\vspace{-0.5cm}
\end{figure*}

Figure~\ref{fig:performancePlots:FailureRate} shows registration failure rate for the following set of pixel sampling rates (in \%): $\{ 0.02, 0.04, 0.065, 0.1, 0.5, 1 \}$. We define a failure as any case with error exceeding 10mm in any of the VOIs. We can see that the proposed approach consistently outperforms other approaches in terms of robustness.

Figure~\ref{fig:performancePlots:mTRE} shows the trimmed mean target registration error (mTRE). We compute the trimmed mTRE as the mTRE of the successful (non-failed) cases. The mTRE is minimal for the proposed approach compared to other methods.  The proposed approach retains high level of accuracy and robustness even with low pixel sampling rates. This allows to significantly reduce computational time in a practical system without exploding the failure rate or reducing accuracy.

Such results support our conjecture that balancing image exploration induced by URS and the exploitation of the prominent image features induced by GMS using a small problem specific training dataset can significantly improve and accelerate the performance of the registration algorithm. Overall, the proposed technique at 0.1\% pixel sampling rate is better than other techniques at 1\% pixel sampling rate, simultaneously maintaining zero failure rate and $1.15$ mm accuracy. Thus on average our approach can use 10 times less pixels for registration, achieve 0 failure rate and improve accuracy over the other two techniques. This is significant improvement over both alternative methods and it allows to reduce the time from 210 seconds per registration for $1\%$ pixel sampling rate to 32 seconds per registration for $0.1\%$ pixel sampling rate in our implementation.

\section{Conclusions and discussion}

In this paper we presented a novel approach to pixel sampling for faster and more accurate registration. Our approach mixes the uniformly random sampling probabilities with those obtained using the gradient magnitude based sampling approach. The mixing parameter that balances image exploration induced by uniform probabilities and the exploitation of image features via gradient magnitude based sampling is learned off-line from a small training dataset. Our experiments with the Vanderbilt RIRE dataset demonstrate that the proposed approach works much faster and produces much more accurate and robust registration results. We conjecture that the concept of mixing the sampling probabilities can be further generalized to obtain even better results. In this case rather than mixing only two sampling methods we could mix three, four or more methods and learn the optimal problem specific mixing coefficients using a small training dataset and a suitable mixing parameter optimization scheme. Exploring this venue based on the experiments with Vanderbilt RIRE dataset and other available datasets as well as testing it on non-rigid registration problems to study the generalizability of the proposed approach seems to be an attractive venue for future research.

\bibliographystyle{splncs03}
\bibliography{WBIR12}

\begin{thebibliography}{10}
\providecommand{\url}[1]{\texttt{#1}}
\providecommand{\urlprefix}{URL }

\bibitem{Benhimane2007}
Benhimane, S., Ladikos, A., Lepetit, V., Navab, N.: Linear and quadratic
  subsets for template-based tracking. In: Proc. IEEE CVPR. Minneapolis, MN,
  USA (Jun 2007)

\bibitem{Brooks2007}
Brooks, R., Arbel, T.: The importance of scale when selecting pixels for image
  registration. In: Proc. CRV. pp. 235--242. Paris, France (May 2007)

\bibitem{Brooks2008}
Brooks, R.: Efficient and Reliable Methods for Direct Parameterized Image
  Registration. Ph.D. thesis, McGill University (2008)

\bibitem{Dellaert1999}
Dellaert, F., Collins, R.: Fast image-based tracking by selective pixel
  integration. In: Proc. ICCV. Kerkyra, Greece (Sep 1999)

\bibitem{Fitzpatrick1998}
Fitzpatrick, J., West, J., Maurer, C.: Predicting error in rigid-body
  point-based registration. IEEE TMI  17(5),  694--702 (Oct 1998)

\bibitem{Freiman2010}
Freiman, M., Werman, M., Joskowicz, L.: A curvelet-based patient-specific prior
  for accurate multi-modal brain image rigid registration. Med. Imag. Anal.
  15(1),  125--132 (Feb 2010)

\bibitem{Kennedy1995}
Kennedy, J., Eberhart, R.: Particle swarm optimization. In: Proc. IEEE Int.
  Conf. Neural Networks. vol.~4, pp. 1942--1948 (nov 1995)

\bibitem{Lu2008}
Lu, X., Zhang, S., Su, H., Chen, Y.: Mutual information-based multimodal image
  registration using a novel joint histogram estimation. Computerized Medical
  Imaging and Graphics  32(3),  202--209 (2008)

\bibitem{Mattes2003}
Mattes, D., Haynor, D., Vesselle, H., Lewellen, T., Eubank, W.: {PET-CT} image
  registration in the chest using free-form deformations. IEEE TMI  22(1),
  120--128 (Jan 2003)

\bibitem{Reeves1995}
Reeves, S.J., Hezar, R.: Selection of observations in magnetic resonance
  spectroscopic imaging. In: Proc. ICIP. vol.~1, pp. 641--644 (Oct 1995)

\bibitem{Sabuncu2004}
Sabuncu, M.R., Ramadge, P.J.: Gradient based nonuniform subsampling for
  information-theoretic alignment methods. In: Proc. IEEE IEMBS. vol.~1, pp.
  1683--1686 (Sep 2004)

\bibitem{Studholme99}
Studholme, C., Hill, D.L.G., Hawkes, D.J.: An overlap invariant entropy measure
  of {3D} medical image alignment. Pattern Recognition  32(1),  71--86 (1999)

\bibitem{Viola1997}
Viola, P., Wells, W.M.: Alignment by maximization of mutual information. IJCV
  24(2),  137--154 (1997)

\bibitem{py06nimg}
Yushkevich, P.A., Piven, J., Hazlett, C.H., Smith, G.R., Ho, S., Gee, J.C.,
  Gerig, G.: User-guided {3D} active contour segmentation of anatomical
  structures: Significantly improved efficiency and reliability. Neuroimage
  31(3),  1116--1128 (2006)

\end{thebibliography}

\end{document}